\documentclass{article}
\usepackage{graphicx}
\usepackage{float}
\usepackage{multirow}
\usepackage{booktabs}
\usepackage{spconf,amsmath,graphicx}
\usepackage{amssymb}
\usepackage{etoolbox,xstring,mfirstuc,textcase}
\usepackage{titlecaps}

\title{\MakeLowercase{\MakeUppercase{M}ulti-Level \MakeUppercase{C}onflict-\MakeUppercase{A}ware \MakeUppercase{N}etwork for \MakeUppercase{M}ulti-Modal \MakeUppercase{S}entiment \MakeUppercase{A}nalysis}}

\name{Yubo Gao$^2$\thanks{Yubo Gao and Haotian Wu contributed equally to this work}, Haotian Wu$^1$, Lei Zhang$^1$\sthanks{Corresponding author. Email: zhlei@bjtu.edu.cn}}
\address{$^1$ Beijing Jiaotong University \\
$^2$ North University of China
}

\begin{document}
\topmargin=0mm

\maketitle
\begin{abstract}
Multimodal Sentiment Analysis (MSA) aims to recognize human emotions by exploiting textual, acoustic, and visual modalities, and thus how to make full use of the interactions between different modalities is a central challenge of MSA. Interaction contains alignment and conflict aspects. Current works mainly emphasize alignment and the inherent differences between unimodal modalities, neglecting the fact that there are also potential conflicts between bimodal combinations. Additionally, multi-task learning-based conflict modeling methods often rely on the unstable generated labels. To address these challenges, we propose a novel multi-level conflict-aware network (MCAN) for multimodal sentiment analysis, which progressively segregates alignment and conflict constituents from unimodal and bimodal representations, and further exploits the conflict constituents with the conflict modeling branch. In the conflict modeling branch, we conduct discrepancy constraints at both the representation and predicted output levels, avoiding dependence on the generated labels. Experimental results on the CMU-MOSI and CMU-MOSEI datasets demonstrate the effectiveness of the proposed MCAN. 
\end{abstract}
\vspace{-0.4cm}
\section{Introduction}
\vspace{-0.2cm}
\label{sec:intro}
In recent years, multimodal sentiment analysis (MSA) has attracted increasingly widespread attention \cite{zadeh2017tensor, tsai2019multimodal, hazarika2020misa, fang2024multi}. Because of the heterogeneity among multimodal data, how to effectively fuse the representations of different modalities and ensure the semantic integrity of modalities is an important research topic in the community of MSA \cite{wang2022cross}. Some of the earlier works focus on the interaction between different modalities on low-level features, which results in limited fusion performance \cite{zadeh2017tensor,liu2018efficient,zadeh2018multi}. Inspired by the attention mechanism's \cite{vaswani2017attention} high-level relationship modeling capabilities, increasing MSA methods introduced attention when fusing unimodal representations. For example, Multimodal transformer (MulT) \cite{tsai2019multimodal} employs the cross-modal attention mechanism to capture multimodal sequence interactions across different time steps. Some other works, such as Text Enhanced Transformer Fusion Network (TETFN) \cite{wang2023tetfn}, Fine-grained Tri-modal Interaction Model (FGTI) \cite{ fang2024multi}, multimodal 3D stereoscopic attention \cite{huang2024multimodal}, etc. have also witnessed the success of the attention-based methods in MSA application.

These methods fuse cross-modal features well but ignore the inherent information and potential conflicts of individual modalities, making the fused information somewhat incomplete. Some studies have noted this problem, either mapping unimodal representations to modality-invariant and modality-specific spaces and modeling them separately subsequently for fusion \cite{hazarika2020misa,yang2022disentangled,li2023decoupled}, or leveraging the multi-task learning (MTL) framework to model inter-modal differences in a supervised learning mode through unimodal label generation \cite{yu2021learning,sun2024novel} or manual annotation \cite{yu2020ch}.

However, these approaches still suffer from some shortcomings. First, there is still a potential conflict between emotional information contained by different bimodal combinations. Considering only inter-unimodal differences is not sufficient. For example, the combination of a smiling expression and a positive word is positive, whereas audio represents sarcasm. In this case, the combination of textual and visual modalities and the combination of textual and acoustic modalities would conflict with the emotional polarity. Secondly, for those methods based on MTL, manual annotation of unimodal labels is costly, whereas label generation methods \cite{yu2021learning,sun2024novel} rely on the quality of unimodal and cross-modal representations, and binary partitioning of the representation center may suffer from insufficient granularity. 

To address these challenges, we propose a multi-level conflict-aware network (MCAN) that models consistency and discrepancy from different levels. Specifically, the MCAN is divided into the main branch and the conflict modeling branch. Wherein, the main branch progressively models the relationship between unimodal and bimodal representations utilizing Micro Multi-step Interaction Network (Micro-MSIN) and Macro Multi-step Intersection Network (Macro-MSIN) and segregates the inter-unimodal and inter-bimodal conflict components hierarchically, then feeds them to the conflict modeling branch. The conflict modeling branch models inter-unimodal and inter-bimodal conflicts through micro conflict-aware cross-attention (Micro-CACA) and macro conflict-aware cross-attention (Macro-CACA), respectively. To avoid introducing unstable representation-based generated labels, the conflict modeling branch directly encourages the unimodal and bimodal representations to generate inconsistent predictions to fully capture the conflict constituents, which will be joint-trained with the main branch. MCAN significantly outperforms the baselines on CMU-MOSI and CMU-MOSEI datasets. Extensive ablation experiments validate the effectiveness of the core component and the influence of the important hyperparameter of MCAN.
\vspace{-0.3cm}

\section{Methodology}
\label{methodology}
\begin{figure*}[htp]
    \centering
    \includegraphics[width=0.9\linewidth]{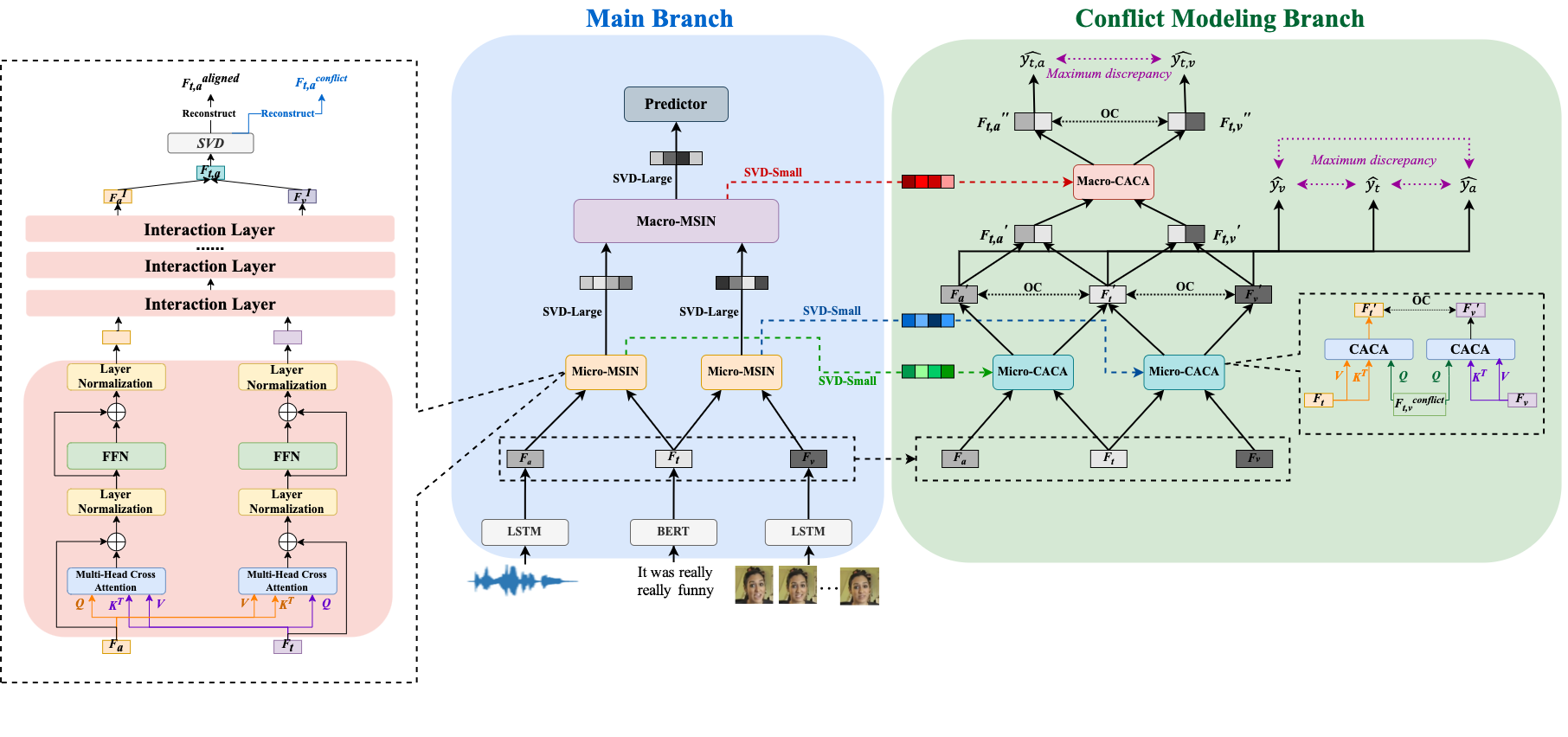}
    \vspace{-0.6cm}
    \caption{The overall framework of MCAN, MSIN and CACA}
    \label{mcan}
\end{figure*}
\vspace{-0.3cm}
The framework of the proposed multi-level conflict-aware network (MCAN) is shown in Figure ~\ref{methodology}. MCAN first conducts feature extraction for the three input modalities. For language modality, we feed the input text into BERT to obtain the language feature $F_t$. While LSTM is adopted to capture the intra-modality interaction $F_v$ and $F_a$ for visual and audio modalities.

\vspace{-0.4cm}
\subsection{Main Branch}
\vspace{-0.2cm}
The function of the main branch is to progressively fuse and align cross-modal representations of different granularities and to segregate conflict constituents. The two core components of the main branch are Transformer-style modules: Micro-MSIN and Macro-MSIN. Micro-MSIN receives $F_{t}$ and $F_{v}$, $F_{t}$ and $F_a$ as inputs, and obtains the outputs $F_{t,a}$ and $F_{t,v}$. Then, inspired by \cite{chen2019transferability,wu2021generalized}, we conduct Singular Value Decomposition (SVD) of $F_{t,a}$ and $F_{t,v}$, and reconstruct the $top-k$ singular values and the corresponding eigenvectors into alignment constituents ($F_{t,a}^{aligned}$ and$F_{t,v}^{aligned}$), which are fed to the Macro-MSIN. The remaining singular values and their corresponding eigenvectors are reconstructed into conflict constituents ($F_{t,a}^{conflict}$, and $F_{t,v}^{conflict}$) to be delivered to the conflict modeling branch.

Macro-MSIN receives $F_{t,a}^{aligned}$ and $F_{t,v}^{aligned}$ as inputs and obtains the fused representation $F_c$, the aligned constituent $F_c^{aligned}$, and the conflicting constituent $F_c^{conflict}$ through a similar computational process to that of Micro-MSIN. The purpose of Macro-MSIN is to fully fuse and align the bimodal representations and separate out the conflict constituents between the bimodal representations. The cascade of Micro-MSIN and Macro-MSIN can make the modeling of MSA modal relationships more adequate and complete. 

\vspace{-0.4cm}
\subsubsection{Micro Multi-step Interaction Network}

The Micro-MSIN modules receive the $F_t$ and $F_a$, $F_t$ and $F_v$ as inputs. Following previous work \cite{yang2024large,sun2024novel,lin2023multi,zhang2023learning,huang2023cross}, we treat the textual modality as the main contributing modality and thus do not set Micro-MSIN between $F_a$ and $F_v$. It consists of multiple layers of Cross-Transformers. Taking the audio-text pairs as an example, the outputs of $(i-1)-th$ layer are $F_t^{(i-1)} \in \mathbb{R}^{n_t \times d}$ and $F_a^{(i-1)} \in \mathbb{R}^{n_a \times d}$, which will be fed to $i-th$ Cross-Transformer layer. For textual modality, $F_t^{(i-1)}$ is transformed into Query to interact with audio modal input features, which are transformed into Key and Value.
The computation for the multi-head cross-modal attention of textual modality is given as follows:
\begin{equation}
 \resizebox{0.85\hsize}{!}{$
\begin{split}
     \operatorname{head}_{j}^{t} = \operatorname{SoftMax}\left(\frac{F_{t}^{(i-1)} W_{Q}\left(F_{a}^{(i-1)} W_{K} \right)^{\top}}{\sqrt{d_{k}}}\right) F_{a}^{(i-1)} W_{V}
\end{split}
$}
\end{equation}
\vspace{-0.2cm}
\begin{equation}
\small
    \text { MultiHead }_{t}=\text { Concat }\left(\text { head }_{1}^{t}, \ldots, \text { head }_{e}^{t}\right) W_{O}
\end{equation}
where $W_Q, W_K, W_V \in \mathbb{R}^{d \times d_k}$, $W_O \in \mathbb{R}^{ed_k \times d}$, $e$ is the number of attention heads. For audio modal, $F_a^{(i-1)}$ will be transformed into a Query and $F_t^{(i-1)}$ will be transformed into Key and Value, then conduct attention computation. Then, the output of cross-modal attention is processed by residual connection, layer normalization and feed-forward neural network (FFN), which is similar to naïve Transformer, and yield output of the $i-th$ interaction layer $F_{g}^{(i)}, g \in {t,a}$.
Assuming that the Micro-MSIN has a total of $I$ layers, the output of the last layer is noted as $F_{t,a}$.
\vspace{-0.2cm}
\begin{equation}
    F_{t,a} = \operatorname{Concatenate}(F_{t}^I, F_{a}^I)
\end{equation}
\vspace{-0.1cm}
To retain the alignment constituents and separate the conflict constituents to the greatest extent possible, we perform SVD, $F_{t,a} = \boldsymbol{U} \boldsymbol{\Sigma } \boldsymbol{V}^{\top} \in \mathbb{R}^{m \times n}$, $\boldsymbol{\Sigma} \in \mathbb{R}^{h \times h}$. In this case, the largest $k$ singular values and the corresponding eigenvectors are considered to be the parts with significant alignment denoted as $F_{t,a}^{aligned}$, while the remaining singular values and the corresponding eigenvectors are regarded as the parts with insignificant alignment, i.e., conflicting, and are denoted as $F_{t,a}^{conflict}$.
\vspace{-0.2cm}
\begin{equation}
\begin{split}
    F_{t,a}^{aligned}&=\boldsymbol{U}_{m \times k} \boldsymbol{\Sigma}_{k \times k} \boldsymbol{V}_{k \times n}^{T} \\
    F_{t,a}^{conflict}&=\boldsymbol{U}_{m \times (h-k)} \boldsymbol{\Sigma}_{(h-k) \times (h-k)} \boldsymbol{V}_{(h-k) \times n}^{T}
\end{split}
\end{equation}
For text-visual pairs, the similar computation process is conducted, which yields $F_{t,v}^{aligned}$ and $F_{t,v}^{conflict}$ as outputs.

\vspace{-0.2cm}
\subsubsection{Macro Multi-step Interaction Network}
\vspace{-0.2cm}
Macro-MSIN serves to model the alignment constituents and conflict constituents between bimodal representations. Macro-MSIN receives $F_{t,a}^{aligned}$ and $F_{t,v}^{aligned}$ as inputs, and its outputs are shown in the following calculations:
\begin{equation}
\small
    Z_{c}^{aligned}, Z_{c}^{conflict} = \operatorname{Macro-MSIN} (F_{t,a}^{aligned}, F_{t,v}^{aligned})
\end{equation}
Micro-MSIN is more fine-grained compared to Macro-MSIN, and they are cascaded to progressively align cross-modal representations at different levels and effectively disentangle conflict knowledge.
\vspace{-0.2cm}
\subsection{Conflict Modeling Branch}
\vspace{-0.2cm}
The conflict modeling branch was designed to receive conflict constituents at different levels from the main branch, and model task conflict in terms of both representations and predicted outputs. It mainly consists of Micro Conflict-aware Cross-Attention (Micro-CACA) and Macro Conflict-aware Cross Attention (Macro-CACA), which are employed for further modeling of conflicts between unimodal representations and bimodal representations, respectively. 
\vspace{-0.2cm}
\subsubsection{Micro Conflict-aware Cross-attention}
The role of Micro-CACA is to adaptively fuse conflict constituents into unimodal representations. To illustrate with the case of text-visual pairs, the conflict constituent $F_{t,v}^{conflict}$ from the main branch will be transformed into Query. The output of the textual modality obtained after Micro-CACA processing is $F_{t}'$
\vspace{-0.2cm}
\begin{equation}
F_{t}'=\operatorname{SoftMax}\left(\frac{F_{t,v}^{conflict}W_Q^c\left(F_{t}W_K^t\right)^{\top}}{\sqrt{d_{c}}}\right) F_{t}W_V^t
\end{equation}
\vspace{-0.2cm}

Similarly, we can obtain Micro-CACA outputs $F_{v}'$ and $F_{a}'$ for visual and acoustic modalities. In particular, the two Micro-CACAs will generate two textual modal representations, which we average as the final outputs. 

To further emphasize the discrepancy between unimodal representations, we impose orthogonal constraints on $F_{t}'$,$F_{v}'$ and $F_{a}'$:
\vspace{-0.2cm}
\begin{equation}
    \mathcal{L}^{oc}_{micro} =  \sum_{p \in\{l, v, a\}} \sum_{q \neq p}\left\|F_{p}^{'^{\top}} F_{q}'\right\|_{F}^{2}
\end{equation}
\vspace{-0.2cm}

Furthermore, we set individual FFN prediction heads for $F_{t}'$,$F_{v}'$ and $F_{a}'$ and encourage them to generate distinct predictions as much as possible to further emphasize the conflicting aspects between unimodal representations at the level of the prediction outputs.
\vspace{-0.2cm}
\begin{equation}
    \mathcal{L}^{diff}_{micro} =  \sum_{p \in\{l, v, a\}} \sum_{q \neq p} \mid \hat{y}_{p}' -\hat{y}_{q}' \mid^{2}
\end{equation}
\vspace{-0.4cm}
\subsubsection{Macro Conflict-aware Cross-attention}
The process of Macro-CACA is similar to that of Micro-CACA. Macro-CACA receives the separated conflict constituents of the main branch Macro-MSIN and transforms them into the Query of cross attention to capture and adaptively fuse inter-bimodal (between $F_{t,a}'$ and $F_{t,v}'$) conflicts. Similarly, the discrepancy constraints at the representation level and the predicted output level of Macro-CACA are represented as follows:
\vspace{-0.2cm}
\begin{equation}
\mathcal{L}^{oc}_{macro} =  \left\|F_{t,v}^{''^{\top}} F_{t,a}''\right\|_{F}^{2}, 
\mathcal{L}^{diff}_{macro} =  \mid \hat{y}_{t,v}'' -\hat{y}_{t,a}'' \mid^{2}
\end{equation}
where $F_{t,v}^{''}$ and $F_{t,a}^{''}$ are features extracted by Macro-CACA, $\hat{y}_{t,v}''$ and $\hat{y}_{t,a}''$ are predicted outputs of $F_{t,v}^{''}$ and $F_{t,a}^{''}$.
The final loss function is represented as follows:
\begin{equation}
\small 
    \mathcal{L} = \mathcal{L}_{main} + \alpha (\mathcal{L}^{oc}_{micro} + \mathcal{L}^{oc}_{macro}) + \beta (\mathcal{L}^{diff}_{micro} + \mathcal{L}^{diff}_{macro})
\end{equation}
where $\mathcal{L}_{main}$ is mean squared error loss, $\alpha$ and $\beta$ are trade-off parameters to control the intensity of conflict modeling.
\vspace{-0.2cm}
\begin{table*}[htp]
\label{tab1}
\centering
\caption{The experiment results on \textbf{CMU-MOSI} and \textbf{CMU-MOSEI} across various evaluation metrics.}
\small
\setlength{\tabcolsep}{4pt} 
\begin{tabular}{lp{1.2cm}p{1.2cm}p{1.2cm}p{1.2cm}p{1.2cm}p{1.2cm}p{1.2cm}p{1.2cm}p{1.2cm}p{1.2cm}}
\toprule
\multirow{2}{*}{Model} & \multicolumn{5}{c}{CMU-MOSI} & \multicolumn{5}{c}{CMU-MOSEI} \\
\cmidrule(lr){2-6} \cmidrule(lr){7-11}
 & Acc2 & Acc7 & F1 & Corr & MAE & Acc2 & Acc7 & F1 & Corr & MAE \\
\midrule
    TFN & 76.8 & 32.5 & 76.3 & 0.601 & 0.998 & 78.5 & 43.7 & 78.0 & 0.665 & 0.709 \\
    LMF & 77.4 & 33.9 & 76.5 & 0.638 & 0.922 & 78.8 & 42.9 & 79.1 & 0.644 & 0.682 \\
    MARN & 78.1 & 34.7 & 77.0 & 0.655 & 0.908 & 79.3 & 44.8 & 79.7 & 0.673 & 0.672 \\
    RAVEN & 79.8 & 36.2 & 79.3 & 0.699 & 0.886 & 80.5 & 45.7 & 80.0 & 0.678 & 0.631 \\
    MulT & 81.3 & 38.4 & 81.4 & 0.734 & 0.802 & 82.9 & 47.7 & 82.8 & 0.744 & 0.586 \\
    MISA & 81.7 & 40.6 & 81.3 & 0.720 & 0.793 & 83.3 & 49.8 & 83.2 & 0.767 & 0.572 \\
    Self-MM & 82.5 & 40.9 & 82.4 & 0.769 & 0.725 & 84.1 & 49.8 & 84.4 & 0.786 & 0.555 \\
    GFML & 83.9 & 41.9 & 83.8 & 0.804 & 0.694 & 85.1 & 50.1 & 84.8 & 0.795 & 0.541 \\
    MMIN & 84.2 & 42.6 & 84.1 & 0.805 & \textbf{0.671} & 85.3 & 50.0 & 85.3 & 0.791 & 0.542 \\
    MSAN & 83.6 & 41.5 & 83.7 & 0.794 & 0.712 & 84.6 & 49.5 & 84.2 & 0.768 & 0.551 \\
    \textbf{MCAN (Ours)} & \textbf{84.5} & \textbf{43.1} & \textbf{84.8} & \textbf{0.811} & 0.675 & \textbf{85.8} & \textbf{51.6} & \textbf{85.9} & \textbf{0.798} & \textbf{0.527} \\
\bottomrule
\end{tabular}
\end{table*}

\section{Experiment}
\label{experiment}
\subsection{Datasets, Metrics and Implementation Details}
We evaluate MCAN on CMU-MOSI \cite{zadeh2016multimodal} and CMU-MOSEI \cite{zadeh2018multimodal} datasets, which are the most widely used benchmark for MSA. Five different metrics are employed to evaluate the performance of MCAN and baselines: binary accuracy (Acc2), 7-class accuracy (Acc7), F1 Score (F1), Pearson correlation (Corr), and mean absolute error (MAE). For the Experimental setting, $\alpha$ and $\beta$ are set to 1e-2 and 1e-3, respectively. Adam is adopted as the optimizer with an initial learning rate 5e-5 for BERT and 1e-4 for other parameters. Additionally, We select the $top-44$ singular values and the corresponding eigenvectors for generating the alignment constituents
\vspace{-0.1cm}
\subsection{Comparison with Baselines}
To validate the effectiveness of our proposed method, the baselines we chose cover classical MSA methods, and recent competitive approaches: \textbf{TFN} \cite{zadeh2017tensor}, \textbf{LMF} \cite{liu2018efficient}, \textbf{MARN} \cite{zadeh2018multi}, \textbf{RAVEN} \cite{wang2019words}, \textbf{MulT} \cite{tsai2019multimodal}, \textbf{MISA} \cite{hazarika2020misa}, \textbf{Self-MM} \cite{yu2021learning},  \textbf{GFML} \cite{sun2024novel}, \textbf{MMIN} \cite{fang2024multi}, \textbf{MSAN} \cite{huang2024multimodal}.

The results of the comparative analysis, as illustrated in Table ~\ref{tab1}, demonstrate that our model achieves significant improvement compared to baselines across different datasets. 
Fusion-based methods such as TFN, and LMF, despite their simplicity, have limited performance due to the difficulty of capturing high-level feature interactions. Compared to these fusion-based methods, attention-based methods such as MARN, RAVEN, and MulT demonstrate improved performance. Benefiting from the excellent high-level relationship capture capabilities of the attention mechanism, MMIN, and MSAN design novel attention modules to fine-grained align the representations of different modalities and achieve performance improvements. Self-MM and GFML focus on the intrinsic differences between modalities by introducing generated labels to model unimodal differences under the MTL framework. In contrast to the above methods, our approach balances alignment and conflict of modal representations at different levels and avoids the introduction of unstable generated labels by encouraging conflicting modeling branches to yield distinct predictions. As a result, the proposed MCAN further improves the performance of MSA.
\vspace{-0.2cm}
\small
\begin{table}[htp]
\label{tab2}
\centering
\caption{Ablation study of MCAN on \textbf{CMU-MOSI}.
“w/o” means without the specific components.}
\begin{tabular}{lccccc}
\hline
\textbf{Ablation} & \textbf{Acc2} & \textbf{Acc7} & \textbf{F1} & \textbf{Corr} & \textbf{MAE}  \\
\hline
\multicolumn{6}{c}{\textbf{Effect of discrepancy constraints}} \\
\hline

w/o $\mathcal{L}_{diff}$ & 82.1 & 42.3 & 82.0 & 0.763 & 0.814 \\
w/o $\mathcal{L}_{oc}$  & 81.9 & 42.2 & 82.0 & 0.759 & 0.816 \\
\hline
\multicolumn{6}{c}{\textbf{Effect of CMB}} \\
\hline
w/o CMB & 82.3 & 42.5 & 82.2 & 0.774 & 0.711 \\
\hline
\multicolumn{6}{c}{\textbf{Effect of truncation positions}} \\
\hline
Top-8 & 79.9 & 36.5 & 80.2 & 0.700 & 0.821 \\
Top-16 & 83.8 & 42.5 & 83.6 & 0.796 & 0.701 \\
Top-24 & 82.5 & 40.5 & 82.6 & 0.745 & 0.771 \\
Top-36 & 84.3 & 42.7 & 84.3 & 0.807 & 0.698 \\
Top-52 & 83.4 & 41.7 & 83.3 & 0.776 & 0.720 \\
Top-64 & 83.0 & 41.1 & 83.0 & 0.762 & 0.742 \\
\hline
\end{tabular}
\end{table}

\subsection{Ablation Study}
The effectiveness of core components and each loss in our method is verified by ablation experiments on the CMU-MOSI dataset, and the results are shown in Table ~\ref{tab2}. We individually removed $\mathcal{L}_{diff}$ and $\mathcal{L}_{oc}$ (Sum of corresponding terms for Micro-CACA and Macro-CACA) to assess the efficacy of these discrepancy constraints constraints. The experimental results reveal that the omission of either $\mathcal{L}_{diff}$ or $\mathcal{L}_{oc}$ results in a noticeable deterioration in model performance. Specifically, $\mathcal{L}_{diff}$ and $\mathcal{L}_{oc}$ function to regularize the feature and prediction aspects, respectively. Furthermore, the experiments verify the effect of the Conflict Modeling Branch (denoted as CMB in Table 2). The design of CMB improves the conflict-capturing ability of our model. Lastly, we confirmed that the choice of the truncation position of singular values in SVD is critical to the outcomes. Different truncation positions will affect the amount of information assigned to the alignment and conflict constituents.

\vspace{-0.4cm}
\section{Conlusion}
\label{conculsion}
In this paper, we develop a novel MCAN for MSA. To balance the discrepancies between unimodal and bimodal representations while fusing and aligning cross-modal representations, MCAN is divided into a main branch and a conflict modeling branch, which are jointly trained in a multi-task learning manner. The former progressively extracts different levels of cross-modal alignment and segregates the conflict constituents through the cascade of Micro-MSIN and Macro-MSIN, while the latter receives these conflict constituents and further models the conflicts. The experimental results show that MCAN outperforms the current state-of-the-art methods. In future work, we will endeavor to further analyze the modal conflict problem at the optimization level (e.g. gradient) and improve the proposed method.

\section{Acknowledge}
\label{acknowledge}
The work was supported by the National Natural Science Foundation 
of China (No.72271017)

\bibliographystyle{IEEEbib}
\bibliography{refs}

\end{document}